\documentclass{article}

\usepackage[final]{exait_2025}

\usepackage{algorithm}
\usepackage[utf8]{inputenc} 
\usepackage[T1]{fontenc}    
\usepackage{hyperref}       
\usepackage{url}            
\usepackage{booktabs}       
\usepackage{amsfonts}       
\usepackage{nicefrac}       
\usepackage{microtype}      
\usepackage{xcolor}         

\usepackage{graphicx}
\usepackage{amsmath}
\usepackage{amssymb}
\usepackage{textcomp}
\usepackage{gensymb}
\usepackage{algorithmic}
\graphicspath{{images/}}
\usepackage{multirow}
\usepackage{arydshln}

\title{Beyond Static Datasets: Robust Offline Policy Optimization via Vetted Synthetic Transitions}

\author{%
  Pedram Agand\thanks{Corresponding author: [firstname][lastname]@gmail.com} \\
    Department of Computing Science,\\
  Simon Fraser University\\
  Burnaby, BC, Canada. \\
    \texttt{pedram\_agand@sfu.ca} \\
    \And
   Mo Chen \\
   Department of Computing Science,\\
   Simon Fraser University \\
   Burnaby, BC, Canada. \\
   \texttt{mochen@cs.sfu.ca} \\
}

\begin{document}

\maketitle

\begin{abstract}
Offline Reinforcement Learning (ORL) holds immense promise for safety-critical domains like industrial robotics, where real-time environmental interaction is often prohibitive. A primary obstacle in ORL remains the distributional shift between the static dataset and the learned policy, which typically mandates high degrees of conservatism that can restrain potential policy improvements. We present MoReBRAC, a model-based framework that addresses this limitation through Uncertainty-Aware latent synthesis. Instead of relying solely on the fixed data, MoReBRAC utilizes a dual-recurrent world model to synthesize high-fidelity transitions that augment the training manifold. To ensure the reliability of this synthetic data, we implement a hierarchical uncertainty pipeline integrating Variational Autoencoder (VAE) manifold detection, model sensitivity analysis, and Monte Carlo (MC) dropout. This multi-layered filtering process guarantees that only transitions residing within high-confidence regions of the learned dynamics are utilized. Our results on D4RL Gym-MuJoCo benchmarks reveal significant performance gains, particularly in ``random'' and ``suboptimal'' data regimes. We further provide insights into the role of the VAE as a geometric anchor and discuss the distributional trade-offs encountered when learning from near-optimal datasets.
\end{abstract}

\section{Introduction}
Offline Reinforcement Learning (ORL) provides a paradigm for extracting high-performance policies from pre-existing, static datasets, effectively bypassing the safety and cost constraints of online exploration \cite{levine2020offline}. While methods like Behavior Cloning (BC) serve as a straightforward baseline, their performance is inherently capped by the quality of the demonstrations within the dataset. To exceed these limits, ORL must stitch together optimal fragments of suboptimal trajectories. However, this process is frequently undermined by the distributional shift problem: when an agent queries actions outside the support of the training data, Q-value overestimation can lead to catastrophic policy failure \cite{fujimoto2019off}.

Existing solutions generally fall into two camps: policy-constrained methods that restrict the agent to familiar actions, and value-pessimism methods that penalize OOD state-action pairs. While effective at stabilizing learning, such measures often result in ``undue conservatism'', where the agent fails to discover superior strategies even when the underlying dynamics would allow for them. Model-based ORL offers a middle ground by learning a simulator of the environment to virtually expand the data horizon. However, the utility of these models is often hindered by the compounding of prediction errors, which can introduce adversarial noise into the training process if left unchecked.

In this work, we propose MoReBRAC (\underline{Mo}del-based \underline{R}estrictive \underline{E}nhanced \underline{B}ehavior-\underline{R}egularized \underline{A}ctor-\underline{C}ritic). MoReBRAC is an uncertainty-aware latent synthesis that constructs a robust world model using a hierarchical LSTM-GRU architecture to generate counterfactual transitions. These transitions are then passed through a \emph{hierarchical uncertainty stack}. In this stack, the VAE filters for global manifold membership, sensitivity checks ensure local stability, and MC Dropout quantifies epistemic uncertainty.

Our primary contributions are threefold: (1) The MoReBRAC framework, which achieves a balance between model-based synthesis and conservative policy regularization; (2) A hierarchical uncertainty pipeline that protects the policy from hallucinated dynamics; (3) An analysis of the VAE’s role as a geometric anchor that prevents manifold drift. Our evaluations demonstrate that MoReBRAC is particularly potent in low-quality data regimes, where it successfully synthesizes the missing ``connective tissue'' required for robust policy improvement.

\section{Related Work}
The landscape of ORL is largely defined by strategies to mitigate distributional shift. Policy-regularized methods maintain proximity to the behavior policy via supervised losses. In parallel, Value-regularized methods adopt a pessimistic view of OOD regions. While these model-free approaches are robust, they are restricted by the ``static'' nature of the data. Model-based ORL attempts to overcome this by learning dynamics models. Our work diverges from these by utilizing a hierarchical uncertainty stack. While ensembles capture model disagreement, our pipeline integrates VAE-based ELBO to capture the ``geographic'' likelihood of the data manifold. This aligns with recent trends in Uncertainty Quantification (UQ), which advocate for multi-modal assessments of model confidence \cite{lakshminarayanan2017simple,padhi2025calibrating}. MoReBRAC synthesizes these threads, using a restrictive world model to augment the conservative foundation of ReBRAC, ensuring that synthetic exploration remains grounded in the true environment manifold. Now we dive deep into each stream of work.
\subsection{Offline Reinforcement Learning (ORL)}
ORL  seeks to learn policies from a static dataset $D = \{ (s_t, a_t, r_t, s_{t+1}) \}_{t=0}^{H-1}$ collected by one or more, potentially unknown, behavior policies $\beta$ \cite{levine2020offline}. This setting is distinct from online RL, where an agent continuously interacts with an environment. The central difficulty in ORL is the \emph{distributional shift} between the behavior policy $\beta$ and the learned policy $\pi$. Standard off-policy RL algorithms often fail in this setting because they may query actions $a \sim \pi(s)$ that are OOD with respect to the dataset, leading to erroneously high Q-value estimates due to function approximation errors and lack of corrective feedback from the environment \cite{fujimoto2019off, kumar2019stabilizing}. This problem is often exacerbated by the bootstrapping nature of temporal difference learning. Numerous approaches have been proposed to tackle distributional shift. These can be broadly categorized:

\textbf{Policy Regularization:} These methods constrain the learned policy $\pi$ to remain close to the state-action visitation distribution of the behavior policy $\beta$. Examples include Behavior Regularized Actor Critic (BRAC) \cite{wu2019behavior}, TD3+BC \cite{fujimoto2021minimalist} which adds a behavior cloning term to the actor loss, and Implicit Q-Learning (IQL) \cite{kostrikov2021offline} which learns an implicit policy by advantage weighting. More specifically, TD3+BC  blends Twin Delayed Deep Deterministic policy gradient (TD3)'s \cite{fujimoto2018addressing} actor-critic framework with a behavioral cloning loss to constrain policy updates, ensuring it remains within the safe bounds of the offline dataset while achieving robust performance in diverse settings.

\textbf{Value Function Regularization:} 
These methods modify the value function learning objective to be pessimistic about OOD actions. Conservative Q-Learning (CQL) \cite{kumar2020conservative} is a prominent example, adding a regularizer that minimizes Q-values for actions sampled from the current policy and maximizes Q-values for actions from the dataset. More recent works like ReBRAC \cite{tarasov2024revisiting} demonstrate that careful architectural and hyperparameter choices can significantly improve the performance of simpler policy regularization methods. Some methods also aim to improve generalization across latent distributions within offline data \cite{wang2025improving}. Recent work also focuses on outcome-driven constraints rather than purely action-based similarity \cite{jiang2025beyond}.

\subsection{Model-Based RL}
Model-based RL methods often learn a model of the environment's dynamics, $\hat{p}(s_{t+1}, r_t | s_t, a_t)$, from data. This learned model can then be used for planning or to generate synthetic data for policy optimization \cite{kidambi2020morel}. In the offline context, model-based approaches are appealing as they can potentially generate more diverse data than what is available in the static dataset, helping to mitigate issues of narrow data coverage \cite{yu2020mopo, kidambi2020morel}.

A key challenge for model-based ORL is that inaccuracies in the learned dynamics model can lead to compounding errors, especially during long rollouts, resulting in unrealistic or even adversarial synthetic data \cite{sims2024edge}. To address this, techniques often involve quantifying model uncertainty and incorporating it into policy optimization. For example, MOPO \cite{yu2020mopo} and MOReL \cite{kidambi2020morel} use ensembles of dynamics models and penalize rewards based on the variance of their predictions, thereby encouraging the policy to stay within regions where the model is confident. COMBO \cite{Yu2021combo} further refines this by training a conservative Q-function on a mix of real and model-generated data. Recent works also explore diffusion models for trajectory generation or data augmentation in offline RL \cite{cao2025model}. The MORE algorithm \cite{zhan2022deepthermal}, which influenced our approach, specifically uses a data-driven simulator for an industrial process and employs checks for generated data. Our work aims to build upon these ideas by integrating a robust suite of uncertainty mechanisms to ensure the quality of augmented data.

\subsection{Uncertainty Quantification in RL}
Uncertainty quantification (UQ) is crucial for reliable decision-making in RL, especially when dealing with limited data or complex environments  \cite{zhu2024uncertainty}. In deep learning, common UQ techniques include \emph{Deep Ensembles} where they train multiple independent models and use the variance of their predictions as an uncertainty measure \cite{lakshminarayanan2017simple}. Another method is \emph{Monte Carlo (MC) Dropout}, where they apply dropout at test time by performing multiple stochastic forward passes to obtain a distribution of predictions, and the variance can serve as an uncertainty estimate \cite{gal2016dropout}. This provides a computationally cheaper approximation to Bayesian inference for large models.
\emph{Variational Autoencoders (VAEs)} \cite{kingma2013auto} are another approach that learn a latent representation of the data. The reconstruction error or the evidence lower bound (ELBO) can be used to detect OOD samples, as data points dissimilar to the training distribution often result in poor reconstructions or lower ELBO values.

\section{The MoReBRAC Framework} 
MoReBRAC is an uncertainty-aware, model-based framework for ORL designed to maximize the utility of static datasets. By synthesizing the conservative policy constraints of ReBRAC \cite{tarasov2024revisiting} with a sophisticated data generation and filtering pipeline inspired by MORE \cite{zhan2022deepthermal}, the framework effectively expands the training manifold. The central mechanism involves generating high-fidelity synthetic transitions through a learned world model; a process defined here as \emph{uncertainty-guided simulated exploration}. This enables the agent to navigate novel state-action sequences in a controlled, virtual environment, thereby mitigating the data scarcity inherent in fixed datasets. The training procedure is bifurcated into two distinct phases: 
\begin{enumerate} 
\item \textbf{World Model Pre-training}: The simulator package, comprising the sequential dynamics model and the VAE, is trained across a diverse aggregation of D4RL datasets (encompassing expert, medium, and random trajectories). This broad exposure prevents the model from overfitting to the bias of a specific behavior policy. Once the model effectively predicts future states and associated rewards, its weights are frozen to provide a stable foundation for the subsequent phase. 
\item \textbf{Policy Optimization}: The frozen simulator generates synthetic trajectories within the main training loop. These transitions are subjected to a multi-stage uncertainty pipeline; only those deemed reliable are integrated into the Prioritized Replay Buffer (PRB). The policy is then refined using the TD3+BC objective, leveraging a hybrid sampling strategy that balances real and synthetic experiences. \end{enumerate}

\subsection{World Model Architecture}
The world model serves as the computational surrogate for the environment, tasked with projecting plausible future states based on historical context. To ensure generalization across diverse dynamics, the model is pre-trained on the full spectrum of available D4RL data.

\subsubsection{Hierarchical State Prediction: LSTM with GRU Refinement}
Capturing the intricate temporal dependencies of sequential decision-making requires an architecture capable of both long-range context retention and high-fidelity transition modeling. We employ a Long Short-Term Memory (LSTM) network as the primary temporal encoder. Taking the previous $n$ states ($n=10$) and the current action as normalized inputs, the LSTM (configured with $l=10$ layers and $h=128$ hidden dimensions) processes the sequence to establish a latent context of the trajectory's history. This historical grounding is vital for capturing underlying momentum or system constraints that single-step models often overlook. Following this initial projection, a Gated Recurrent Unit (GRU) module, inspired by \cite{agand2024dmfuser}, refines the predicted transition. This GRU transition engine, utilizing a hidden dimension $h_g=512$ and a prediction horizon $\tau=5$, updates the state sequence auto-regressively. By decoupling the task into history encoding (LSTM) and iterative refinement (GRU), MoReBRAC maintains structural stability during multi-step rollouts, effectively suppressing the accumulation of prediction errors over extended horizons.

\subsubsection{VAE-based Manifold Encoding}
To define the geometric boundaries of the offline data, a Variational Autoencoder (VAE) is trained in parallel. This module employs a U-Net-inspired architecture with $h_v=750$ hidden dimensions and a latent dimension $z$ set to twice the action dimension. By minimizing the Evidence Lower Bound (ELBO) loss, comprising reconstruction accuracy and KL-divergence, the VAE learns to map the high-dimensional state-action distribution. During data synthesis, it serves as a ``topological anchor,'' providing a quantitative measure of how closely a generated state aligns with the original training manifold.

\subsection{Hierarchical Uncertainty Quantification}
The integrity of the simulated exploration process depends on the rigorous filtering of candidate transitions. MoReBRAC utilizes a hierarchical uncertainty stack to evaluate transitions across three distinct failure modes, ensuring that only in-distribution, stable, and high-confidence samples reach the replay buffer.

\textbf{Manifold Support (VAE ELBO):} The pre-trained VAE assesses whether a synthetic transition $(s,a)$ resides within the support of the behavior policy. Transitions yielding an ELBO value $p(s,a)$ below a threshold $E_t$ are categorized as out-of-distribution (OOD) and immediately discarded. This geometric filter prevents the agent from learning from ``hallucinated'' dynamics in regions of the state space where the world model lacks empirical grounding.

\textbf{Local Prediction Stability (Sensitivity Analysis):} To identify regions where the dynamics model exhibits brittle behavior, we introduce perturbations to the input states and actions. The variance in the resulting state predictions is measured; if this sensitivity exceeds a threshold $S_t$, the transition is deemed unstable. This check ensures that the synthesis remains locally robust and resistant to numerical noise.

\textbf{Epistemic Confidence (MC Dropout):} 
We quantify model uncertainty arising from data sparsity using Monte Carlo (MC) dropout. By performing $k_\text{mc}=3$ stochastic forward passes with varied dropout masks, we compute the variation among the predicted next states. A variation exceeding threshold $D_t$ indicates low epistemic confidence, steering the exploration away from regions where the model’s knowledge is insufficient.

\subsection{Restrictive Exploration and Reward Shaping}
During trajectory synthesis, transitions are evaluated sequentially. Any failure in the hierarchical uncertainty stack results in the immediate disposal of the transition. If the VAE ELBO exceeds a maximum tolerance $E_\text{max}$, indicating a terminal deviation from the data manifold, the entire rollout is truncated to prevent the propagation of erroneous signals. To further temper optimism for transitions that remain near the manifold's edge, MoReBRAC applies an uncertainty-aware reward penalization. The modified reward $R_\text{new}(s,a)$ is computed as:
\begin{equation}
R_\text{new}(s,a) = \frac{R(s,a)}{1+K(l_p-p(s,a))},
\label{eq:rp}
\end{equation}
where $R(s,a)$ is the simulator's predicted reward, $p(s,a)$ is the ELBO loss, $l_p$ is the lower bound threshold, and $K$ is a tuning parameter. This scaling ensures that even ``valid'' synthetic data is treated with appropriate skepticism, guiding the policy toward high-confidence trajectories.

\subsection{Hybrid Prioritized Replay and Policy Learning}
Validated, reward-penalized transitions are integrated into a Prioritized Replay Buffer (PRB) partitioned into three segments: (1) static offline data, (2) filtered synthetic data, and (3) prioritized high/low reward transitions. To balance grounding in reality with the benefits of synthesis, we utilize a hybrid sampling curriculum. During the \textbf{Warm-up Phase}, sampling is restricted to the offline dataset to establish a robust policy foundation. In the subsequent \textbf{Hybrid Phase}, the mini-batch distribution shifts to a mixture (e.g., 60\% offline, 30\% synthetic, and 10\% prioritized salient transitions). This curriculum ensures the agent retains the stability of the original dataset while selectively exploiting the diversity of the augmented manifold. Final policy updates are performed using the ReBRAC-inspired actor-critic framework, allowing MoReBRAC to achieve a more comprehensive understanding of the environmental dynamics than possible through model-free methods alone.

\section{Experiments and Results}

\subsection{Experimental Setup}
We evaluate MoReBRAC on standard Gym-MuJoCo continuous control tasks using datasets from the D4RL benchmark \cite{fu2020d4rl}, specifically focusing on `halfcheetah`, `hopper`, and `walker2d` environments across `random`, `medium`, `medium-replay`, `medium-expert`, and `expert` dataset types. Our comparisons include several strong ensemble-free baselines: ReBRAC \cite{tarasov2024revisiting}, MORE \cite{zhan2022deepthermal} (adapted with an LSTM-only model for its simulator for fair comparison in this context), TD3+BC \cite{fujimoto2021minimalist}, IQL \cite{kostrikov2021offline}, and SAC-RND \cite{nikulin2023anti}.

Architectural hyperparameters for the policy component are primarily derived from the established ReBRAC and TD3+BC configurations. Performance is reported as normalized scores averaged across five independent random seeds using the final training checkpoint. This multi-seed approach captures the variance inherent in high-dimensional control tasks while ensuring the statistical significance of our findings.

\subsection{Performance Evaluation}
The performance of MoReBRAC compared to baselines on D4RL Gym-MuJoCo tasks is presented in Table \ref{table:main}.  The mean-wise best results among algorithms are highlighted with \textbf{bold}, and the second best performance is \underline{underlined}. Our approach demonstrates robust performance across various dataset qualities, highlighting the benefits of its uncertainty-aware simulated exploration.

On \textbf{Random Datasets}, which contain noisy and largely suboptimal trajectories, MoReBRAC achieves the highest average score (22.97), significantly outperforming ReBRAC (18.67) and MORE (16.37). This performance suggests that our synthetic synthesis pipeline effectively acts as a connective tissue, generating the necessary transitions to stitch together suboptimal fragments. The notable gain in \texttt{hopper-random} (19.1) versus ReBRAC (8.1) indicates that our multi-modal filters successfully prune misleading signals, allowing the agent to focus on physically plausible, high-reward sequences within the latent space.

In \textbf{Medium and Full-Replay Datasets}, representing more structured but still suboptimal data, MoReBRAC (83.39) performs comparably to ReBRAC (83.37) and outperforms MORE (72.53). The benefits of simulated exploration are present, but less pronounced than in random data, as the quality of the base data set is higher. The benefits here are driven by the world model’s ability to refine the existing trajectories through local augmentation. This helps the policy navigate around state-action regions where the behavior policy was inconsistent but the underlying dynamics are stable. 

\begin{figure}[t]
\begin{center}
\centerline{\includegraphics[width=0.8\linewidth,trim={0cm  0cm 0cm 0cm },clip]{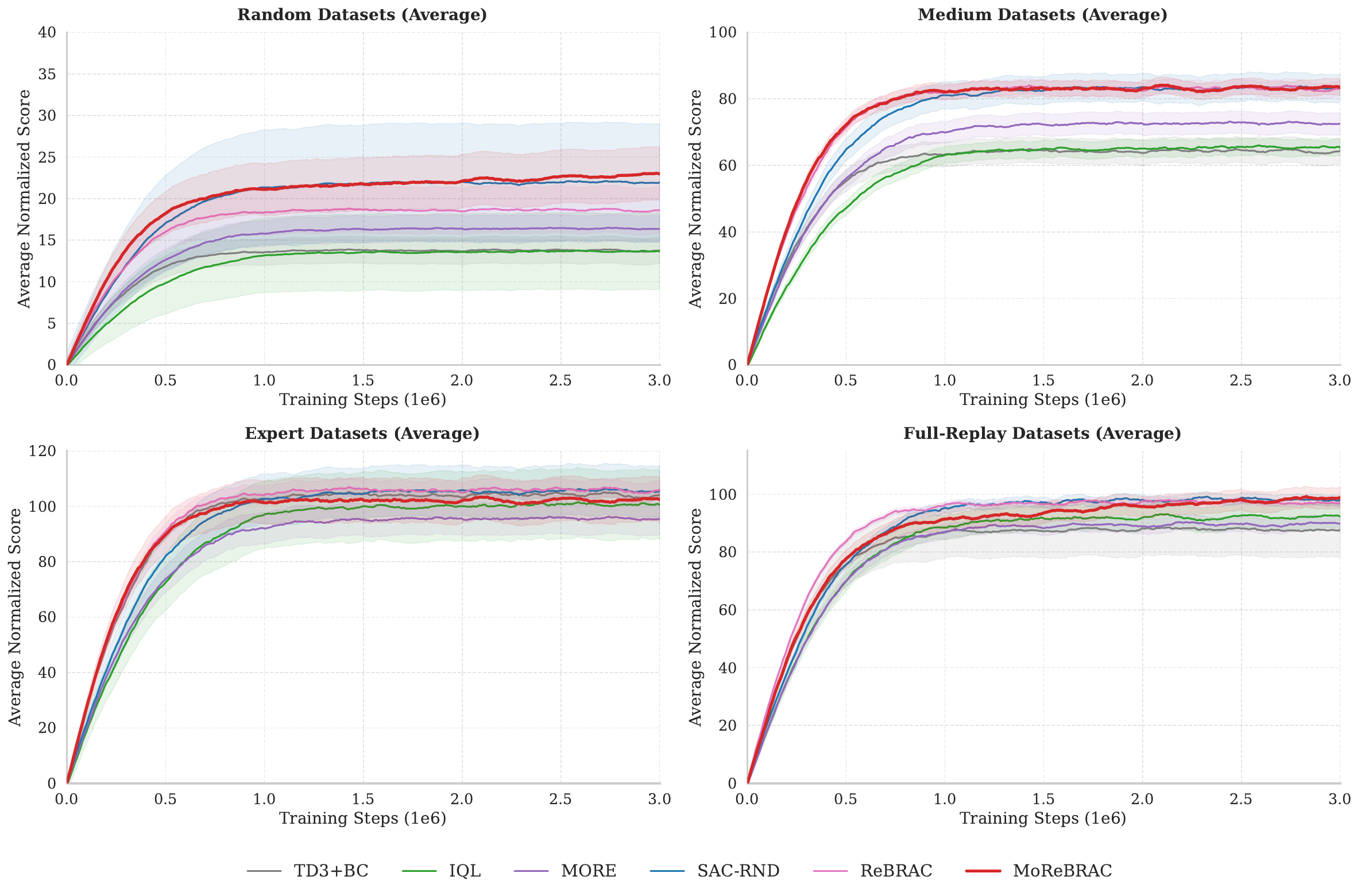}}
\caption{Average Normalized Scores across D4RL Datasets and halfcheetah, hopper, and walker2d.}
 \label{fig:MoReBRAC_profile}
\end{center}
\vskip -0.2in
\end{figure}

A slight performance divergence is observed in \textbf{Expert Datasets}, where MoReBRAC (102.37) is marginally outperformed by the pure ReBRAC baseline (106.1). This outcome aligns with our design priorities. In near-optimal data regimes, the state-action manifold is exceptionally narrow. Introducing synthetic transitions, even those rigorously vetted, can lead to a \emph{distributional dilution} effect. Augmenting an already optimal dataset creates a mixture distribution, analogous to a medium-expert setting, which can introduce slight variance into the Actor-Critic updates. In these scenarios, the framework's inherent conservative bias and reward penalization prioritize robustness and safety over aggressive imitation, a trade-off often required in practical robotic applications to avoid overfitting to a single, narrow demonstration. The curves in Fig. \ref{fig:MoReBRAC_profile} illustrate the training progression of MoReBRAC against state-of-the-art baselines across five random seeds. In Random and Full-Replay regimes (top-left, bottom-right), MoReBRAC demonstrates superior long-term learning, leveraging uncertainty-vetted synthetic data to continue improving beyond the plateau of baselines like ReBRAC. In the Medium regime (top-right), performance is competitive, indicating that synthetic augmentation does not hinder learning in structured datasets. The Expert regime (bottom-left) reveals a slight performance trade-off; MoReBRAC's conservative updates result in a stable but slightly lower plateau compared to pure behavior cloning methods, highlighting its focus on robust, uncertainty-aware policy iteration over aggressive imitation of a narrow expert manifold.

\begin{table}[ht]
\centering
\caption{Performance comparison across tasks. Average normalized score over the final evaluation on Gym-MuJoCo tasks. The symbol ± represents the standard deviation across the seeds.}
\label{table:main}
\scalebox{0.75}{
\begin{tabular}{ll|ccccc|c}
\hline
\textbf{Type} & \textbf{Task Name} & \textbf{TD3+BC} & \textbf{IQL} & \textbf{MORE} & \textbf{SAC-RND} & \textbf{ReBRAC}  & \textbf{MoReBRAC}\\ \hline
\multirow{4}{*}{Random}&halfcheetah & 30.9 ± 0.4 & 19.5 ± 0.8 & 27.1 ± 2.4 & 27.6 ± 2.1 & \underline{29.5 ± 1.5}& \textbf{30.9 ± 3.2}\\
&hopper & 8.5 ± 0.7 & 10.1 ± 5.9 & 12.2 ± 1.2 & \textbf{19.6 ± 12.4} & 8.1 ± 2.4 & \underline{19.1 ± 3.7}\\
&walker2d&2.0 ± 3.6 & 11.3 ± 7.0 & 9.8 ± 1.3 & \underline{18.7 ± 6.9} & 18.4 ± 4.5& \textbf{18.9 ± 2.7}\\
&avg&13.8&13.63&16.37&\underline{21.97}&18.67&\textbf{22.97} \\ \hline
\multirow{4}{*}{Medium}&halfcheetah  & 54.7 ± 0.9 & 50.0 ± 0.2 & 48.9 ± 1.5 & \underline{66.4 ± 1.4} & 65.6 ± 1.0& \textbf{66.8 ± 1.2}\\ 
&hopper & 60.9 ± 7.6 & 65.2 ± 4.2 & 71.5 ± 5.8 & 91.1 ± 10.1 & \textbf{102.0 ± 1.0 }&\underline{100.3 ± 1.8}\\
&walker2d & 77.7 ± 2.9 & 80.7 ± 3.4 & 74.8 ± 2.8 & \textbf{92.7 ± 1.2} & 82.5 ± 3.6 &\underline{83.07 ± 4.3}\\
&avg&64.43&65.3&72.53&\textbf{83.4}&83.37&\underline{83.39} \\ \hline
\multirow{4}{*}{Expert}&halfcheetah &93.4 ± 0.4 & 95.5 ± 2.1 & 89.3 ± 2.1 & \underline{102.6 ± 4.2} & \textbf{105.9 ± 1.7}& 93.7 ± 3.2\\ 
&hopper & \underline{109.6 ± 3.7} & 108.8 ± 3.1 & 97.4 ± 11.5 & \textbf{109.8 ± 0.5} & 100.1 ± 8.3& 102.7 ± 12.4 \\
&walker2d & \underline{110.0 ± 0.6} & 96.9 ± 32.3 & 99.57 ± 3.7 & 104.5 ± 22.8 & \textbf{112.3 ± 0.2}& 110.7 ± 8.4\\ 
&avg&104.33&100.4&95.42&\textbf{105.63}&\underline{106.1}&102.37 \\ \hline
\multirow{4}{*}{Full-replay}&halfcheetah &75.0 ± 2.5 & 75.0 ± 0.7 & 79.4 ± 2.8 & 81.2 ± 1.3 & \underline{82.1 ± 1.1} &\textbf{84.3 ± 4.2}\\
&hopper & 97.9 ± 17.5 & 104.4 ± 10.8 & 98.9 ± 1.6 & \underline{107.4 ± 0.8} & 107.1 ± 0.4&\textbf{108.2 ± 2.5} \\ 
&walker2d & 90.3 ± 5.4 & 97.5 ± 1.4 & 90.7 ± 3.4 & \textbf{105.3 ± 3.2} & 102.2 ± 1.7&\underline{104.2 ± 4.1}  \\ 
&avg&87.73&92.3&89.67&\underline{97.97}&97.13&\textbf{98.9} \\ \hline
&\textbf{Score} & 67.57 & 67.91 & 68.50 & \textbf{77.24} & 76.32& \underline{76.90}\\ \hline
\end{tabular}}
\end{table}

\subsection{Ablation Studies}
To understand the contribution of each key component in MoReBRAC, we conducted ablation studies, presented in Table \ref{table:abl}. Each component was disabled while keeping others active, and results are reported as the mean of average normalized scores over three seeds.

\textbf{The VAE as a Global Manifold Anchor:} 
The most critical finding is the 61\% performance collapse observed when the VAE is removed (\textit{No VAE}). This confirms the VAE's role as more than a simple filter; it acts as a global geometric anchor. While the LSTM and GRU predict local dynamics, the VAE ensures the ``geographic'' validity of the state. Without this constraint, the synthetic rollouts suffer from compounding manifold drift, where the simulator begins to generate transitions in ``data vacuums'' that have no support in the true environment dynamics.

\textbf{Epistemic and Local Stability:} 
The omission of MC Dropout (\textit{No MC}) and the sensitivity check (\textit{No sensitivity}) resulted in drops of 6\% and 2\%, respectively. These mechanisms provide a nuanced assessment of model confidence. While the VAE catches macroscopic out-of-distribution (OOD) errors, MC Dropout captures epistemic uncertainty, the model's ``internal confusion'' due to data sparsity. The sensitivity check further ensures that synthesis occurs only in locally stable regions, preventing the policy from relying on brittle, high-variance predictions.

\textbf{Pessimistic Reward Synthesis:} 
Disabling the reward penalization (\textit{No penalize reward}) caused a 13\% decline in performance. This highlights the necessity of treating synthetic data with a degree of structural skepticism. By scaling rewards according to the VAE’s ELBO loss, MoReBRAC implements a form of Pessimistic Value Estimation. This discourages the agent from becoming overly optimistic about synthesized paths near the manifold boundary, thereby ensuring that policy updates remain grounded in high-certainty data.

Finally, the PRB and hybrid sampling components contribute to the stabilization of the learning curriculum. By grounding the policy in the real offline distribution during the warm-up phase before introducing synthetic diversity, the framework avoids early-training divergence and ensures a more consistent convergence across varied task types.

\begin{table}[ht]
\centering
\caption{Ablation study for design choices: each modification was disabled while keeping all the others. For brevity, we report the mean of average normalized scores over three unseen training seeds. }
\label{table:abl}
\scalebox{0.9}{
\begin{tabular}{l|cccc}
\toprule
\textbf{Ablation} & \textbf{halfcheetah} & \textbf{hopper} & \textbf{walker2d} & \textbf{avg} \\ 
\hline
No VAE & 31.57 (-54\%) & 34.17 (-59\%) & 23.17 (-71\%)& 29.64 (-61\%) \\
No MC& 61.23 (-11\%) & 76.24 (-8\%) & 80.25 (+1\%) & 72.57 (-6\%) \\
No sensitivity& 67.38 (-2\%)  & 82.24 (-0.2\%)& 76.41 (-4\%) & 75.34  (-2\%)\\
No penalize reward& 61.42 (-11\%)  & 71.32 (-14\%)  & 67.25 (-15\%)  & 66.66  (-13\%) \\
No PRB& 69.78 (+1\%)& 80.47 (-3\%)  & 78.21 (-1\%) & 76.15  (-1\%) \\
No hybrid sampling& 65.24 (-5\%) & 78.67 (-5\%) & 81.23 (+3\%) & 75.05 (-2\%) \\
\hline
MoReBRAC & 68.93  & 82.58 & 79.22 & 76.91 \\
\end{tabular}}
\end{table}

\section{Conclusion}
This paper introduced MoReBRAC, an ORL framework that leverages uncertainty-aware latent synthesis to expand the boundaries of static datasets. By integrating a hierarchical LSTM-GRU world model with a principled, multi-stage uncertainty quantification pipeline, we demonstrate that it is possible to generate reliable synthetic trajectories that enhance policy learning. Our evaluations on the D4RL benchmark highlight the framework's particular strength in suboptimal and random data regimes, where it successfully balances the exploitation of known data with a cautious, uncertainty-vetted expansion of the state-action manifold. Ultimately, MoReBRAC offers a robust pathway for moving beyond the limits of static observations, providing a scalable solution for high-performance, safety-conscious ORL.

\newpage
\appendix

\section{Supplementary Material}
The proposed method is shown in Algorithm \ref{alg:proposed}, where $R(s,a)$ is the transition reward, $l_p$ is the lower bound threshold, $K$ is a tuning parameter, and $p(s,a)$ is the ELBO loss of the generated trajectory.

\begin{algorithm}[h]
\caption{Training and Evaluation Process for MoReBRAC}
\begin{algorithmic}[1]
\STATE \emph{~~~Stage 1: Train the Simulator Package}
\STATE Load the D4RL dataset of different variations.
\STATE Sort the trajectories based on length to enable mini-batch sampling.
\REPEAT
\STATE Choose a random index and create a mini-batch of sliding windows.
\STATE Forward path for LSTM + GRU model to predict future states. 
\STATE Compute the reward given the predicted state and the MSE loss.
\STATE Backpropage losses for both LSTM and GRU
\STATE Evaluate on held-out trajectories using $R^2$ values for both states and rewards.
\UNTIL{Validation loss increases more than maximum patience. }
\STATE Train the VAE model to minimize ELBO loss on validation loss \cite{pinheiro2021variational}.
\STATE \emph{~~~Stage 2: Policy Update}
\STATE \textbf{- Step 1: Filter Generated Data based on uncertainty checks}
\STATE Reset the simulated environment
\REPEAT
\STATE 1. Compute the ELBO for each transition using the VAE model.  If it exceeds the threshold ($E_t$), discard the transition.
\STATE 2. Apply perturbations to the input states and actions, measuring the variance in output. If it exceeds the threshold ($S_t$), discard the transition.
\STATE 3. Apply three MC runs for different random dropout, compute the variation. Discard the transition if the variation is above threshold ($D_t$). 
\STATE Apply reward penalization based on the VAE loss as in Eq. (\ref{eq:rp})
\UNTIL{If the ELBO exceeds a maximum tolerance ($E_\text{max}$), truncate the trajectory and go to Step 1.}
\STATE \textbf{- Step 2: Prioritized Replay Buffer}
\STATE Add generated data to the PRB generated portion.
\STATE Separate high/low reward transitions for PRB prioritized portion.
\STATE \textbf{- Step 3: Hybrid sampling}
\IF {Warm-up (early epochs)}
\STATE Sample purely from offline portion of PRB.
\ELSE 
\STATE Sample 60\% from offline, 30\% from generated, and 10\% from prioritized portion.
\ENDIF
\STATE \textbf{- Step 4: Train Actor-Critic networks}
\STATE Update policy using the ReBRAC method \cite{tarasov2024revisiting}.
\STATE \textbf{- Step 5: Evaluate}
\STATE Evaluate the policy on the average of simulator several trajectories.
\STATE Fine-tune the model using gradient clipping, learning rate scheduling, and early stopping according to validation performance.
\STATE \emph{~~~Stage 3: Test the model}
\STATE Evaluate the model on the actual environment.
\end{algorithmic}
\label{alg:proposed}
\end{algorithm}

\section{Discussion}
MoReBRAC establishes a framework for \emph{uncertainty-guided latent synthesis} within the ORL paradigm. By using a dual-recurrent world model to generate synthetic transitions and subsequently vetting them through a hierarchical uncertainty stack, the framework transcends the static boundaries of the initial dataset. This approach directly addresses the ``undue conservatism'' often found in ORL, where agents are penalized for any deviation from the behavior policy. Instead of treating all out-of-sample actions as equally not useful, MoReBRAC identifies ``trust regions'' where the world model exhibits high epistemic and geometric confidence. This allows for a form of simulated trajectory stitching, enabling the agent to navigate novel state-action sequences that are physically plausible, without accessing to the real-world interactions. Our empirical findings demonstrate that this synthesis is particularly vital in ``low-signal'' regimes datasets characterized by noise or sparse coverage of high-reward regions. In these scenarios, the world model acts as an interpolator, filling the gaps between disjointed trajectories and providing the connective tissue necessary for policy improvement.

\begin{figure}[t]
\begin{center}
\centerline{\includegraphics[width=0.5\linewidth,trim={1cm  0cm 2.2cm 0.5cm },clip]{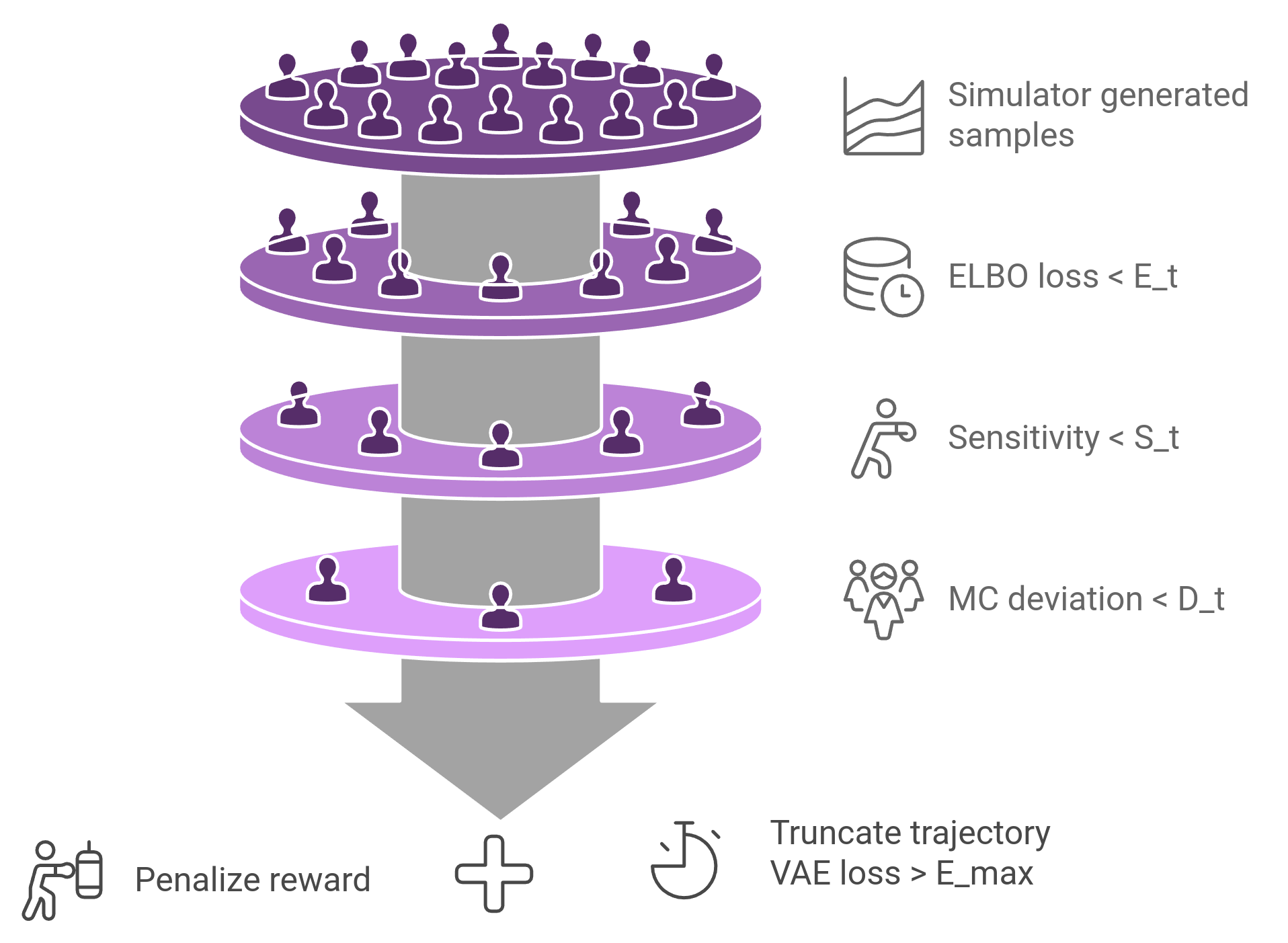}}
\caption{MoReBRAC restrictive exploration}
 \label{fig:MoReBRAC}
\end{center}
\vskip -0.2in
\end{figure}

The architecture of MoReBRAC reflects broader trends in reliable artificial intelligence. The reliance on a hierarchical LSTM-GRU simulator engages with the necessity of capturing long-range temporal dependencies to ensure rollout stability \cite{yu2020mopo}. Furthermore, the integration of multi-modal uncertainty, visualized as a restrictive filter in Fig. \ref{fig:MoReBRAC}, is essential for the deployment of RL in safety-critical domains. In environments such as robotic control or autonomous navigation, unverified exploration is often catastrophic. MoReBRAC’s pipeline ensures that synthetic exploration remains productive and grounded, providing a safer alternative to the ``trial-and-error'' nature of traditional online RL.

The capacity for uncertainty-aware data generation carries significant implications for policy generalizability. By selectively augmenting sparsely represented but critical regions of the state space, MoReBRAC effectively creates new learning signals. This mechanism shares conceptual foundations with self-supervised learning, where auxiliary data is generated to refine the model's internal representation of environmental dynamics \cite{liu2021self}. Moreover, the prioritization of high-reward transitions within the PRB, coupled with reward penalization for manifold boundaries, aligns with the principles of active learning \cite{brame2016active}. Beyond performance gains, the presence of a vetted simulator facilitates crucial pre-deployment validation. This enables extensive hyperparameter optimization and stress-testing in a simulated environment before any physical interaction occurs. For safety-critical systems, this layer of simulated verification is a prerequisite for real-world adoption, offering a bridge between the theoretical potential of ORL and its practical implementation.

\subsection{Limitations and Future Directions}
While MoReBRAC exhibits strong performance, its efficacy remains dependent on the fidelity of the world model. Model-based RL is perpetually susceptible to the ``compounding error'' problem, where minor inaccuracies at step $t$ propagate into catastrophic hallucinations at following steps. While our hierarchical filters mitigate this, the exploration achieved is fundamentally a sophisticated interpolation within the learned data manifold. If the initial dataset fails to capture a fundamental environmental mechanic, no amount of synthetic synthesis can recover that information \cite{fu2020d4rl, zhan2022offline}. Furthermore, as discussed in the ``distributional dilution'' effect for our framework's conservative bias, while essential for safety, can result in slight performance ceilings in purely expert datasets. In these cases, the rigorous filtering and reward penalization prioritize manifold support over aggressive imitation, which may be suboptimal for pure behavior cloning tasks. Finally, the computational complexity of maintaining the Uncertainty Stack (VAE, MC Dropout, and Sensitivity analysis) introduces a training overhead that may be less desirable for simple tasks where model-free methods suffice. Future work could investigate more efficient proxies for these uncertainty measures to reduce the time of the training loop.


{\small
\bibliographystyle{plainnat}
\bibliography{ref}
}


\end{document}